\documentclass[letterpaper, 10 pt, conference]{ieeeconf}  

\IEEEoverridecommandlockouts                              

\usepackage{amsmath} 
\usepackage{amssymb}  
\usepackage{multirow}
\usepackage{algorithmic}
\usepackage{algorithm}
\usepackage{graphicx}
\usepackage{subcaption}

\title{\LARGE \bf
Robust Deep Reinforcement Learning in Robotics via Adaptive Gradient-Masked Adversarial Attacks
}

\author{
    Zongyuan Zhang\textsuperscript{\rm 1},
    Tianyang Duan\textsuperscript{\rm 1},
    Zheng Lin\textsuperscript{\rm 2},
    Dong Huang\textsuperscript{\rm 1}, 
    Zihan Fang\textsuperscript{\rm 2},
    Zekai Sun\textsuperscript{\rm 1}, \\
    Ling Xiong\textsuperscript{\rm 3},
    Hongbin Liang\textsuperscript{\rm 4},
    Heming Cui\textsuperscript{\rm 1}, 
    Yong Cui\textsuperscript{\rm 5},
    Yue Gao\textsuperscript{\rm 2}
    \thanks{$^{1}$ Department of Computer Science, The University of Hong Kong, Hong Kong, China.}
    \thanks{$^{2}$ School of Computer Science, Fudan University, Shanghai, China.}
    \thanks{$^{3}$ School of Computer and Software Engineering, Xihua University, Chengdu, China.}
    \thanks{$^{4}$ School of Transportation and Logistics, Southwest Jiaotong University, Chengdu, China.}
    \thanks{$^{5}$ Department of Computer Science and Technology, Tsinghua University, Beijing, China.}
}

\begin{document}

\maketitle
\thispagestyle{empty}
\pagestyle{empty}

\begin{abstract}
Deep reinforcement learning (DRL) has emerged as a promising approach for robotic control, but its real-world deployment remains challenging due to its vulnerability to environmental perturbations. Existing white-box adversarial attack methods, adapted from supervised learning, fail to effectively target DRL agents as they overlook temporal dynamics and indiscriminately perturb all state dimensions, limiting their impact on long-term rewards. To address these challenges, we propose the Adaptive Gradient-Masked Reinforcement (AGMR) Attack, a white-box attack method that combines DRL with a gradient-based soft masking mechanism to dynamically identify critical state dimensions and optimize adversarial policies. AGMR selectively allocates perturbations to the most impactful state features and incorporates a dynamic adjustment mechanism to balance exploration and exploitation during training. Extensive experiments demonstrate that AGMR outperforms state-of-the-art adversarial attack methods in degrading the performance of the victim agent and enhances the victim agent's robustness through adversarial defense mechanisms.


\end{abstract}

\section{INTRODUCTION}
Robotic systems are increasingly deployed in mobile and distributed applications, ranging from autonomous navigation \cite{wei2024autonomous,lin2022channel,hu2024toward, 10135137,lin2024adaptsfl,hu2024agentscodriver,song2023emma}, intelligent transportation \cite{10323097,lin2022tracking, khalil2024advanced,lin2023pushing,fang2024ic3m,lin2022v2i}, and industrial manufacturing \cite{han2023survey, 10100908}.  Traditional robotic control methods have demonstrated success in structured environments with predefined tasks but face limitations in dynamic scenarios, uncertainty handling, and experiential learning. Deep reinforcement learning (DRL) has emerged as a viable alternative for robotic control \cite{lin2021softgym}. Unlike manually designed rule-based approaches, DRL enables agents to optimize behaviors through trial-and-error interactions with their environment. By learning policies that map states to actions to maximize long-term rewards, DRL excels in complex tasks characterized by delayed feedback and temporal dependencies.

The robustness of DRL policies is critical for real-world robotic applications, as DRL agents are highly sensitive to environmental perturbations \cite{shi2024distributionally,yuan2024itpatch}. Small input variations caused by sensor noise, environmental changes, or adversarial attacks can disrupt decision-making and lead to catastrophic failures. White-box adversarial attacks are effective for assessing DRL robustness by identifying vulnerabilities in learned policies. With full access to model architecture and parameters, such attacks systematically generate perturbations to evaluate policy networks \cite{schott2024robust}. Adversarial training using these perturbations enables DRL agents to improve resilience to state variations, ensuring reliable performance in real-world scenarios.

However, existing white-box attack methods face significant challenges when targeting DRL agents, as they primarily rely on local gradient information to generate perturbations \cite{duan2021advdrop, chen2024diffusion, chen2024content}. These methods, adapted from supervised learning, assume temporal independence and focus on instantaneous state-action mappings, overlooking the temporal dynamics of Markov Decision Processes (MDPs). Consequently, they fail to generate perturbations that effectively disrupt cumulative rewards over extended time horizons. Additionally, these methods indiscriminately apply perturbations across all states without identifying critical features that impact performance. This is particularly problematic in high-dimensional state spaces, where only a subset of variables—such as specific joint angles in robotic control—are essential for policy execution. While some approach weight perturbations based on policy gradient magnitudes \cite{oikarinen2021robust, hickling2023robust, haydari2021adversarial}, they do not align with the core attack objective of minimizing cumulative rewards, as agents can adapt by selecting alternative actions to maintain comparable long-term performance.

To address these challenges, we propose \textbf{A}daptive \textbf{G}radient-\textbf{M}asked \textbf{R}einforcement (AGMR) attack, a white-box method that integrates DRL with a gradient-based soft masking mechanism to dynamically identify critical state dimensions and optimize adversarial policies. AGMR introduces a soft mask function to allocate perturbations selectively across state dimensions, focusing on features that have the greatest impact on the victim agent's decision-making. Furthermore, AGMR incorporates a dynamic adjustment mechanism for the interpolation factor in the soft mask function, enabling the adversarial agent to balance exploration and exploitation during training. By leveraging gradient magnitudes to quantify the importance of state dimensions, AGMR adjusts its attack strategy in response to the evolving dynamics of the environment and task. Experimental results show that the proposed AGMR method enhances the effectiveness of adversarial attacks, and consistently outperforms state-of-the-art adversarial attack methods across several key metrics, including reward reduction, velocity reduction, and an increase in the number of falls. Additionally, AGMR demonstrates the ability to improve the robustness of the victim agent through adversarial defense mechanisms.

\section{RELATED WORK}
Adversarial attacks expose the vulnerabilities of deep neural networks (DNNs) by introducing carefully crafted perturbations into input data, leading to incorrect predictions during inference~\cite{ZW_TCOM_2024,wang2024ultralola}. These perturbations, though imperceptible to humans, can significantly alter DNN outputs \cite{akhtar2021advances}. White-box attacks constitute a critical category of adversarial attacks, where the adversary has full access to the model, including its architecture, parameters, and gradients. Fast Gradient Sign Method (FGSM) \cite{goodfellow2014explaining} is a seminal white-box attack that efficiently generates adversarial perturbations by leveraging the gradient of the loss function with respect to the input, addressing the computational inefficiencies of earlier approaches. Projected Gradient Descent (PGD) \cite{madry2017towards} enhances attack effectiveness through iterative gradient-based updates, projecting perturbed inputs back into the constrained space after each step. Wong et al. \cite{wong2020fast} improved attack efficiency by introducing random initialization points in FGSM-based attacks. Schwinn et al. \cite{schwinn2023exploring} increased attack diversity by injecting noise into the output while mitigating gradient obfuscation caused by low-confidence predictions. Beyond standard white-box attacks, various techniques have been proposed to improve adversarial transferability \cite{wang2021admix, dong2018boosting}. These include random input transformations \cite{xie2019improving}, translation-invariant perturbation aggregation \cite{dong2019evading}, and substituting momentum-based gradient updates with Nesterov accelerated gradients \cite{lin2019nesterov}.

Adversarial attacks in DRL have been widely studied, revealing critical vulnerabilities in agent policies \cite{ilahi2021challenges}. Huang et al. \cite{huang2017adversarial} demonstrated that policy-based DRL agents are highly susceptible to adversarial perturbations on state observations, showing that FGSM attacks can significantly degrade performance in Atari 2600 games. Pattanaik et al. \cite{pattanaik2017robust} introduced adversarial examples by computing gradients of the critic network with respect to states and integrated them into the training of Deep Double Q-Network (DDQN) and Deep Deterministic Policy Gradient (DDPG), enhancing robustness. Lin et al. \cite{lin2017tactics} proposed strategically timed attacks that selectively perturbed key decision-making states, achieving high attack success rates with minimal perturbations. Recent work has shifted towards theoretical modeling of adversarial attacks within the Markov Decision Process framework. Weng et al. \cite{weng2019toward} introduced a systematic evaluation framework for DRL robustness in continuous control, defining two primary threat models: observation manipulations and action manipulations. Zhang et al. \cite{zhang2020robust} proposed SA-MDP, which provides a theoretical foundation for modeling state adversarial attacks within MDPs. Oikarinen et al. \cite{oikarinen2021robust} developed RADIAL-RL, a general framework for training DRL agents to enhance resilience against adversarial attacks, and introduced Greedy Worst-Case Reward as a new evaluation metric for agent robustness.

\section{METHODOLOGY}
\subsection{Problem Formalization}
DRL is formalized as Markov Decision Process (MDP), which is defined as a tuple $\left \langle \mathcal{S},\mathcal{A},R,\mathcal{P},\gamma \right \rangle $, where $\mathcal{S}$ and $\mathcal{A}$ denote the state and action spaces, and $\gamma$ denotes the discount factor. At each time step $t$, the agent selects an action $a_t$ from its policy $\mu \left ( \cdot \mid s_t \right )$. The environment returns a reward $R\left ( s_t,a_t \right )$ and transitions to the next state $s_{t+1}$ according to the state-transition probability function $\mathcal{P} \left ( \cdot \mid s_t,a_t \right )$. The state value function is defined as $V \left ( s_t \right ) = \mathbb{E}_{a\sim \mu,s\sim \mathcal{P} } \left [ {\textstyle \sum_{k=0}^{\infty } \gamma ^k R\left ( s_{t+k},a_{t+k} \right )}  \right ]  $ which represents the expected discounted return starting from state \( s_t \) under policy \( \mu \). Similarly, the action-value function (Q-function) that quantifies the expected discounted return for selecting action \( a_t \) in state \( s_t \) under policy \( \mu \) is given by  $Q^{\mu } \left ( s_t,a_t \right ) = R\left ( s_t,a_t \right )+\mathbb{E}_{s_{t+1}\sim \mathcal{P} } \left [ V^{\mu } \left ( s_{t+1} \right )   \right ]  $. The agent's objective is to learn a policy $\mu$ that maximizes the expected discounted return:
\begin{equation}
\label{eq:1}
\begin{aligned}
J\left ( \mu  \right ) = {\textstyle \sum_{t=0}^{\infty }\gamma ^t\mathbb{E}_{a\sim \mu,s\sim \mathcal{P} } \left [ R\left ( s_t,a_t \right )  \right ] }.
\end{aligned}
\end{equation}

In the remainder of the paper, we assume all Q-functions and policies are parameterized. For brevity, we denote \( Q^\mu _\phi \) as \( Q^\mu \) and the policy \( \mu _\varphi \) as \( \mu \), where \( \varphi \) represents the policy network parameters. We call policy $\mu \left ( \cdot \mid s \right ) $ as the victim policy (i.e., the policy under attack) and assume it has converged in the environment.

White-box adversarial attacks, commonly used to expose vulnerabilities in deep neural networks (DNNs) through gradient-based perturbations, naturally apply to victim policy networks in DRL \cite{schott2024robust}. To be Specific, given a victim policy $\mu \left ( \cdot \mid s \right ) $, the objective of an adversarial attack is to introduce a minimal perturbation $\eta$ to input state $s$, such that the victim policy network outputs an action that maximally deviates from the original action $a$. This can be formulated as a constrained optimization problem:
\begin{equation}
\label{eq:3}
{\mathrm{arg} \max}_{s^*} J\left ( s^*,a \right ) ,\quad \mathrm{s.t.} \ \left \| \eta \right \| _p\le \epsilon,
\end{equation}
where $s^*=s+ \eta $ represents the perturbed state, $\eta$ denotes the adversarial perturbation, $J\left ( \cdot ,\cdot  \right ) $ is the loss function, typically mean squared error or cross-entropy. The constraint $\ \left \| \eta \right \| _p\le \epsilon$ ensures that the perturbation magnitude remains within a predefined threshold $\epsilon\in \left ( 0,\infty  \right ) $, where $\ \left \| \cdot  \right \| _p$ denotes the $L_p$ norm. 

\subsection{Adversarial Policy Based on Reinforcement Learning}
While white-box adversarial attacks have shown effectiveness in supervised learning, their direct adaptation to DRL faces significant challenges due to the temporal and sequential nature of MDPs. These methods rely on local gradient information to generate adversarial perturbations, assuming temporal independence and focusing solely on instantaneous state-action mappings. Such assumptions overlook the long-term effects of perturbations on trajectory evolution, making gradient-based perturbations ineffective over extended time horizons. Agents can often adapt by leveraging alternative actions to mitigate short-term performance degradation, thereby diminishing the attack’s impact over time. Moreover, indiscriminate application of perturbations across all state dimensions fails to exploit the local importance of features, particularly in high-dimensional spaces, where only a subset of variables—such as critical joint angles in robotic tasks—significantly impacts policy execution. In such cases, targeted perturbations on critical dimensions can lead to significant trajectory divergence or fundamental behavioral shifts, effects that existing methods fail to effectively capture.

To address these challenges, our aim is to develop a white-box adversarial attack method based on the DRL framework, which autonomously identifies and exploits vulnerable state dimensions in the victim agent's policy while considering the impact on long-term rewards. Let $\nu \left ( \cdot \mid s,\mu  \right ) $ denote the adversarial policy (i.e., adversarial agent's policy), representing the attacker's strategy under a white-box setting, where the attacker has full access to the victim's policy $\mu$ and generates adversarial perturbations based on both the current state $s$ and $\mu$. Since the victim agent follows a fixed policy $\mu$ throughout the attacked rollout, it can be regarded as part of the environment’s dynamics from the adversarial agent’s perspective. The adversarial policy samples a perturbation $\eta _t\sim \nu \left ( \cdot \mid s_t,\mu \right ) $ at each time step $t$, which is then applied to the victim agent’s observation. The victim agent selects an action according to its policy
\begin{equation}
a _t \sim \mu  \left ( \cdot \mid s_t+ \eta_t \right ),
\end{equation}
where $s_t+ \eta_t$ is the perturbed state.The single attack process is independent of the environment dynamics, forming a one-step sequential decision process. This interaction between the adversarial policy and the victim policy unfolds over multiple time steps, resulting in an attacked rollout that represents the sequential effects of adversarial perturbations on the victim agent's behavior. Formally, the rollout can be expressed as:
\begin{equation*}
s_0\overset{\nu  }{\rightarrow} \eta _0 \overset{\mu }{\rightarrow} a_0\overset{\mathcal{P}  }{\rightarrow}s_1\overset{\mu }{\rightarrow}\dots \overset{\mathcal{P} }{\rightarrow}s_t \overset{\nu  }{\rightarrow} \eta _t \overset{\mu }{\rightarrow}a_t\overset{\mathcal{P}  }{\rightarrow}\dots
\end{equation*}

\begin{figure}[t]
  \centering
  \includegraphics[width=0.48\textwidth]{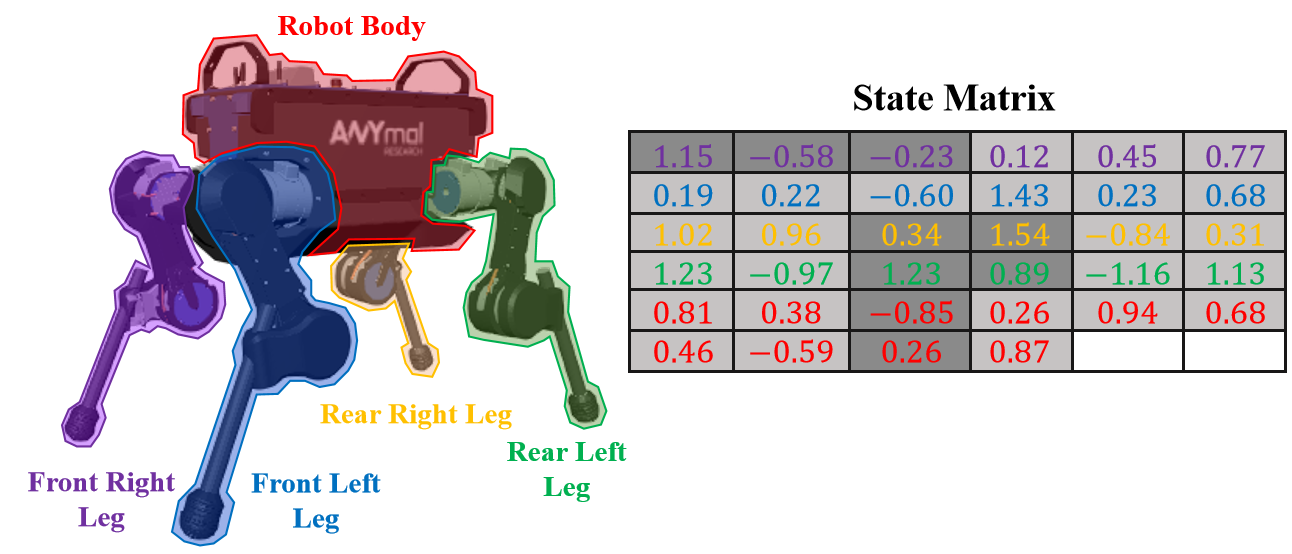}
  \caption{Illustration of AGMR's soft-masked attack mechanism. \textbf{Left}: The robotic system is decomposed into five components (the body and four legs). \textbf{Right}: The heatmap of the state matrix visualizes perturbation magnitudes across state dimensions.}
  \label{fig:1}
\end{figure}

In contrast to existing white-box attack methods described by Eq.~\ref{eq:3}, the adversarial agent aims to degrade the victim agent's performance by minimizing its expected return while satisfying perturbation constraints:
\begin{equation}
\label{eq:objective}
\small
J\left ( \nu \right ) = \min_{\nu} { \sum_{t=0}^{\infty }\gamma ^t\mathbb{E}_{a\sim \mu,\eta \sim \nu,s\sim \mathcal{P} } \left [ R\left ( s_t,a_t \right ) \right ] },\ \mathrm{s.t.} \ \left \| \eta \right \| _p\le \epsilon,
\end{equation}
where $R\left ( s_t,a_t \right )$ represents the reward function and $\gamma$ is the discount factor. To this end, we propose \textbf{A}daptive \textbf{G}radient-\textbf{M}asked \textbf{R}einforcement (AGMR) attack, a white-box method based on DRL to identify critical state dimensions via a gradient-based soft masking mechanism and optimize adversarial policies with adaptive-magnitude perturbations. AGMR employs a soft mask function, defined as $M_{\mathrm{soft}}(s) = \beta M(s) + (1-\beta)(1-M(s))$, where $M :\mathcal{S} \to \left \{ 0,1 \right \} $ is a binary mask function used to identify critical state dimensions, and $\beta \in (0,1)$ is an interpolation factor that balances the emphasis between critical and redundant dimensions. The perturbed state generated by the adversarial policy is represented as
\begin{equation}
\label{eq:6}
\nu \left ( \cdot \mid s,\mu  \right ) =\epsilon \cdot M_{\mathrm{soft}}(s) \cdot \mathrm{sign}\left ( \nabla _s J \left ( s',a \right )  \right ) ,
\end{equation}
where $s'= s+\varepsilon \cdot \mathcal{N} \left ( \mathbf{0},\mathbf{I}   \right ) $, $a\sim \mu \left ( \cdot \mid s\right ) $, $\varepsilon \in \left ( 0,1 \right ] $ is the scaling factor,  and $ \mathcal{N} \left ( \mathbf{0},\mathbf{I}   \right )$ represents a multivariate standard normal distribution that introduces small perturbations to prevent gradient vanishing. As shown in Figure~\ref{fig:1}, the soft-masked attack mechanism enables AGMR to focus on critical state dimensions. AGMR generates dimension-specific perturbations, enabling targeted manipulation of state information corresponding to critical robotic components.

\subsection{Automatic Dynamic Adjustment of Interpolation Factor}
In the previous section, we proposed AGMR, a white-box adversarial attack method based on DRL that allocates attack magnitudes across state dimensions using a soft mask function. Selecting an appropriate interpolation factor is critical for ensuring attack effectiveness. However, during the early stages of adversarial agent training, the soft mask function may incorrectly identify critical state dimensions due to insufficient learning. This misidentification results in inefficient perturbation allocation, degrading attack performance and hindering the exploration of optimal attack policies. At this stage, a smaller interpolation factor is required to allow the adversarial agent to flexibly explore various perturbation directions. As training progresses, a larger interpolation factor becomes necessary to focus attacks on critical state dimensions, thereby imposing stronger interference on the victim policy.

Furthermore, the importance of state dimension features may shift significantly due to changes in the optimization of the adversarial policy. State dimensions that are critical at certain stages may become less important over time, while previously redundant dimensions may gain importance. These dynamic changes render a fixed interpolation factor inadequate for adapting to different training phases. Consequently, a mechanism is needed to automatically adjust the interpolation factor in response to task and environmental dynamics. Manual tuning of the interpolation factor is impractical, as different tasks and environments demand varying configurations, and the evolving importance of state dimensions is difficult to anticipate.

We propose a gradient-magnitude-based dynamic adjustment mechanism for the interpolation factor, leveraging the insight that state dimensions with larger gradient magnitudes have a greater impact on the victim agent's decision-making process. Specifically, we define the gradient magnitude as the gradient of the objective function $J\left ( s',a \right )$ denoted by Eq.~\ref{eq:6} with respect to the state $s$:
\begin{equation}
\label{eq:7}
g=\nabla _sJ\left ( s',a \right ).
\end{equation}

By utilizing the binary mask $M(s)$ in the adversarial agent, the gradient of the objective function can be decomposed into critical and redundant components:
\begin{equation}
\label{eq:8}
g_{\text{critical}} = M(s) \odot g, \ g_{\text{redundant}} = (1-M(s)) \odot g,
\end{equation}
where $\odot$ denotes element-wise multiplication. To quantify the impact of critical and redundant dimensions, we compute the $L_p$ norm of the gradient magnitudes, normalized by the number of corresponding dimensions:
\begin{equation}
\label{eq:9}
\bar{g}_{\text{critical}} = \frac{\left \| g_{\text{critical}} \right \|_p }{\left \| M(s) \right \|_p }, \ \bar{g}_{\text{redundant}} = \frac{\left \| g_{\text{redundant}} \right \|_p }{\left \| 1-M(s) \right \|_p}.
\end{equation}

The interpolation factor $\beta$ is dynamically adjusted based on the relative magnitudes of the critical and redundant gradients:
\begin{equation}
\label{eq:10}
\beta = \sigma \left ( \frac{\bar{g}_{\text{critical}}}{\bar{g}_{\text{critical}}+\bar{g}_{\text{redundant}}}  \right ),
\end{equation}
where $\sigma \left ( \cdot  \right ) $ is the sigmoid function to ensure that $\beta \in \left ( 0,1 \right ) $. When the critical dimensions dominate (i.e., $\bar{g}_{\text{critical}} \gg \bar{g}_{\text{redundant}}$), $\beta$ approaches $1$, prioritizing perturbations on critical dimensions. Conversely, when redundant dimensions have comparable or larger gradient magnitudes, $\beta$ decreases, enabling broader exploration of state dimensions or mitigating the risk of overfitting to specific state features that may lose relevance as the adversarial policy evolves.

\subsection{On-Policy Training Scheme for AGMR}

\begin{algorithm}[t]
\footnotesize
\caption{AGMR training algorithm}
\label{alg:1}
\begin{flushleft}
\textbf{Input}: victim agent's policy $\mu \left ( \cdot \mid s \right ) $, batch size $N$, discount factor $\gamma $
\end{flushleft}
\begin{algorithmic}[1] 
\STATE {\bf{Random initialization:}} mask function $M(\cdot )$ with $\theta^M$ and value function $V\left ( \cdot  \right ) $ with $\theta^V$, and replay buffer $\mathcal{D}$
\FOR{each episode}
\STATE Initialize state $s_0$ and $T\gets 0$
\FOR{time step $t$}
\STATE $\eta _t \sim \nu \left ( \cdot \mid s_t,\mu  \right ) $
\STATE $a_t \sim \mu \left( \cdot \mid s_t + \eta_t \right )$
\STATE Execute $a_t$, compute reward $r_t=R\left ( s_t,a_t \right ) $, and store transition $\left ( s_t,a_t,r_t,s_{t+1} \right ) $ in $\mathcal{D}$.
\IF{$s_{t+1}$  is terminal}
\STATE $T\gets t+1$
\STATE Calculate $\widehat{R} _t$ and store return in $\mathcal{D}$:
\begin{center} 
$\widehat{R} _t = \sum_{k=0}^{T-t-1} \gamma ^k r_{t+k}+\gamma^{T-t} V\left ( s_T  \right )$
\end{center}
\STATE  Calculate $\widehat{A} _t$ and store advantage in $\mathcal{D}$:

\begin{center} 
$\widehat{A} _t= \widehat{R}_t -V\left ( s_t  \right )$
\end{center}
\ENDIF
\ENDFOR

\FOR{each epoch}
\STATE Sample a batch of  $\left ( s_i,a_i,r_i,s_{i+1},\widehat{R}_i,\widehat{A}_i   \right ) $ from $\mathcal{D}$
\STATE Update the value function $V\left ( \cdot  \right ) $ by minimizing the loss:

\begin{center}
$\mathcal{L} \left ( V \right ) =\frac{1}{N}\sum_{i=0}^{N} \left [ \hat{R}_i -V\left ( s_i \right )  \right ] ^2$
\end{center}
\STATE Update the mask function $M(\cdot )$ by minimizing the objective:

\begin{center}
$J\left ( \nu  \right )  =\frac{1}{N}\sum_{i=0}^{N}  \hat{A}_i$ 
\end{center}
\ENDFOR
\ENDFOR
\end{algorithmic}
\end{algorithm}

To effectively train the AGMR adversarial attack algorithm, we adopt an on-policy RL framework combined with Generalized Advantage Estimation (GAE) \cite{schulman2015high}. On-policy framework exhibits heightened sensitivity to minor variations in the victim agent's behavioral patterns. In contrast to off-policy methods, which may suffer from distribution shifts due to the utilization of outdated experiences, the on-policy framework ensures that policy updates are consistently derived from the most recent attack-victim interactions. This approach enables enhanced real-time adaptation of perturbation strategies while maintaining update stability through the exclusive use of on-distribution samples from current policy trajectories. Specifically, the advantage function $\widehat{A}_t$, which measures how much better an action is compared to the expected return under the current policy, is computed at time step $t$ as:
\begin{equation}
\widehat{A} _t = \sum_{k=0}^{T-t-1} \left ( \gamma \lambda  \right ) ^k \delta _{t+k},  
\end{equation}
where $\delta _t = r_t+\gamma V\left ( s_{t+1}  \right )-V\left ( s_t  \right )$. We set $\lambda = 1$ to enable GAE to fully accumulate discounted rewards while incorporating value function estimates, thereby providing an unbiased estimation of the advantage function. This setup is particularly suitable for tasks with long-term dependencies, such as adversarial attacks in robotic manipulation, as it captures the complete impact of future rewards without the need for additional weighting or truncation of temporal difference errors. Additionally, AGMR employs a parameterized value function to approximate the expected return for a given state under the current policy. The value function $V$ is updated by minimizing the following mean squared error loss:
\begin{equation}
\mathcal{L} \left ( V \right ) =\mathbb{E} _{ s_i,\hat{R}_i \sim \mathcal{D}  }\left [ \hat{R}_i -V\left ( s_i \right )  \right ] ^2,
\end{equation}
where $\widehat{R} _i = \sum_{k=0}^{T-i-1} \gamma ^k r_{i+k}+\gamma^{T-i} V\left ( s_T  \right )$ denotes the expected return, and $\mathcal{D}$ denotes the replay buffer of on-policy trajectories sampled from the victim policy under attack $\mu \left ( \cdot \mid s+\eta  \right ) $. The mask function $M(s)$, similar to the victim policy $\mu(\cdot \mid s)$ and the value function $V(s)$, is parameterized by a neural network $\theta^M$. The adversarial policy is optimized by minimizing the objective function represented by Eq.~\ref{eq:objective}.

The AGMR training algorithm is detailed in Algorithm~\ref{alg:1}. The training process begins by initializing the mask function $M(\cdot)$ with network parameters $\theta^M$, the value function $V(\cdot)$ with network parameters $\theta^V$ and a replay buffer $\mathcal{D}$ to store transitions (line 1). For each episode, the initial state $s_0$ is set, and the episode length counter $T$ is initialized to zero (line 3). During each time step of the episode, the perturbed state is generated based on the current state $s_t$ and victim policy $\mu$, after which the victim agent selects an action $a_t$ according to the perturbed state (lines 5-6). The selected action $a_t$ is executed in the environment, yielding a reward $r_t$ and transitioning to the next state $s_{t+1}$. The transition tuple $\left(s_t, a_t, r_t, s_{t+1}\right)$ is stored in the replay buffer $\mathcal{D}$ for subsequent training (line 7). If the environment reaches a terminal state $s_{t+1}$, the episode length $T$ is updated. Returns $\widehat{R}_t$ and advantages $\widehat{A}_t$ for all time steps $t$ in the episode are computed and stored in the replay buffer $\mathcal{D}$ (lines 9–11). Training occurs after the collection of trajectories. For each training epoch, a batch of $N$ transitions, returns, and advantages is sampled from the replay buffer $\mathcal{D}$ (line 15). The value function $V$ is updated by minimizing the mean squared error between the predicted value and the stored returns $\widehat{R}i$ (line 16). The mask function $M(\cdot)$, defined in the adversarial policy in Eq.~\ref{eq:6}, is optimized by minimizing the objective $J(\nu)$ (line 17).

\begin{figure}[t]
    \centering
    \includegraphics[width=1\linewidth]{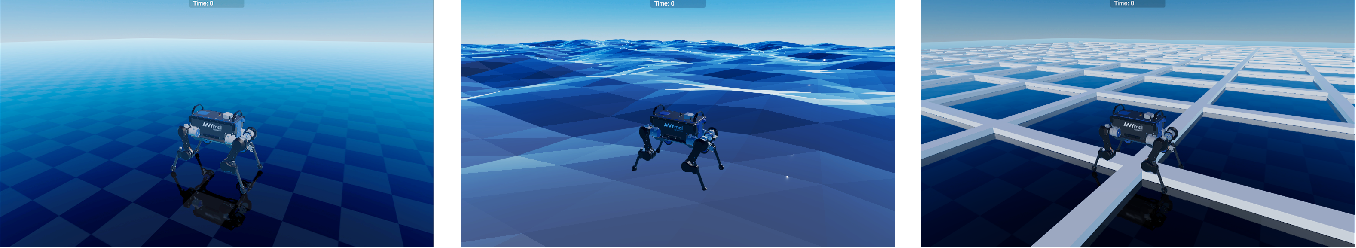}
    \caption{Visualization of the robotic agent's locomotion task across three distinct terrains: flat (left), hill (middle), and obstacles (right).}
    \label{fig:raisim}
\end{figure}

\begin{table}[]
\renewcommand{\arraystretch}{1.2}
\centering
\caption{Hyperparameters}
\begin{tabular}{ll}
\hline
\multicolumn{2}{c}{Victim Agent Hyperparameters} \\
\hline
Victim Policy Network & $2\times128$ FC layers \\
Victim Value Network & $2\times128$ FC layers \\
Clipping parameter & 0.2 \\
Discount factor & 0.998 \\
GAE parameter & 0.95 \\
Initial learning rate & $5\times10^{-4}$ \\
\hline
\multicolumn{2}{c}{Adversarial Agent Hyperparameters} \\
\hline
Adversarial Policy Network & $3\times64$ FC layers \\
Adversarial Value Network & $3\times64$  FC layers \\
Learning rate & $3\times10^{-4}$ \\
\hline
\end{tabular}
\label{tab:hyper}
\end{table}

\section{EVALUATION}

\begin{figure}[htbp]
\centering
\begin{subfigure}[t]{0.49\linewidth}
\centering
\includegraphics[height=8.cm]{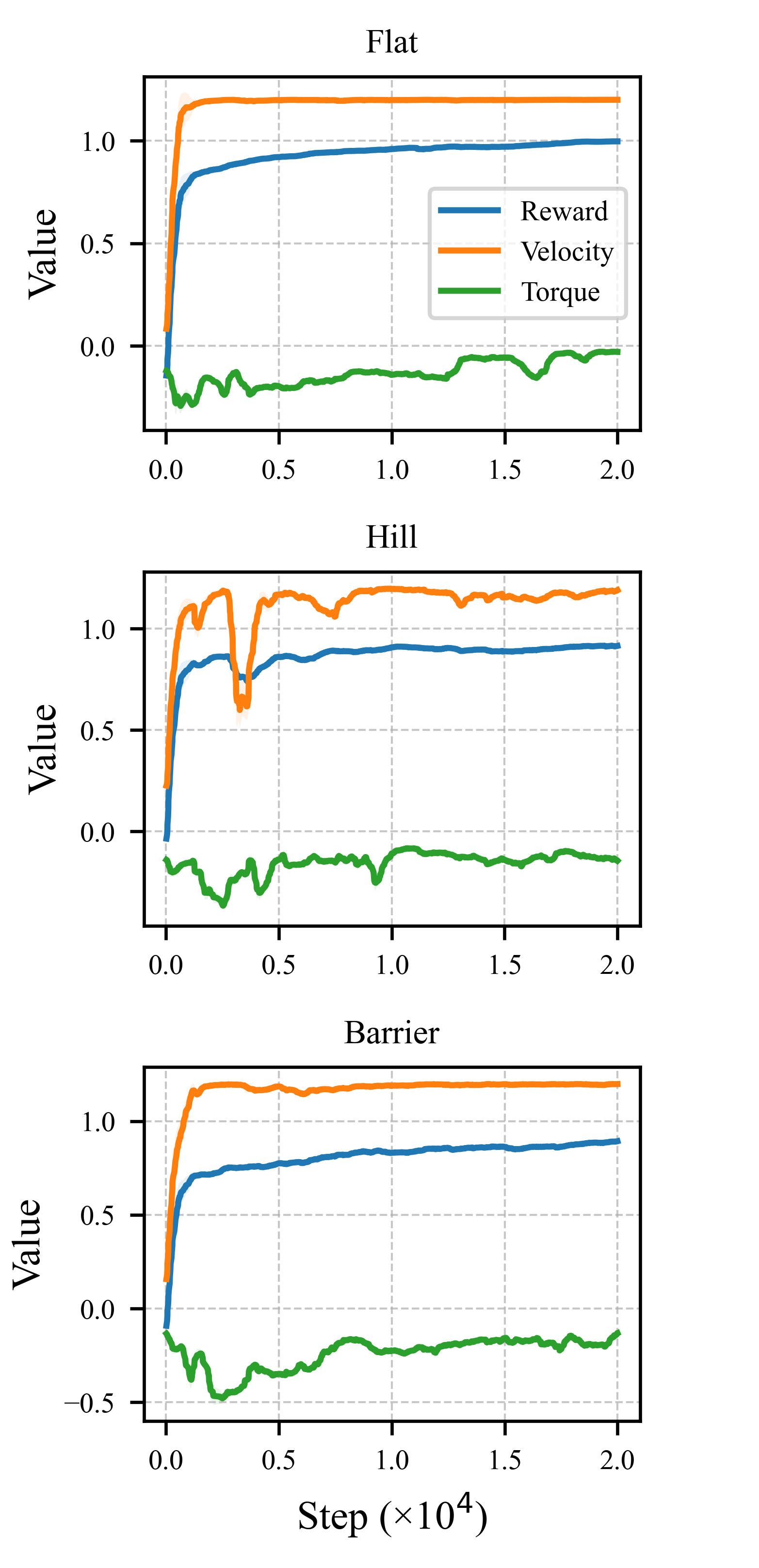}
\caption{}
\label{fig:victim_train}
\end{subfigure}
\kern-0.5cm 
\begin{subfigure}[t]{0.49\linewidth}
\centering
\includegraphics[height=8.cm]{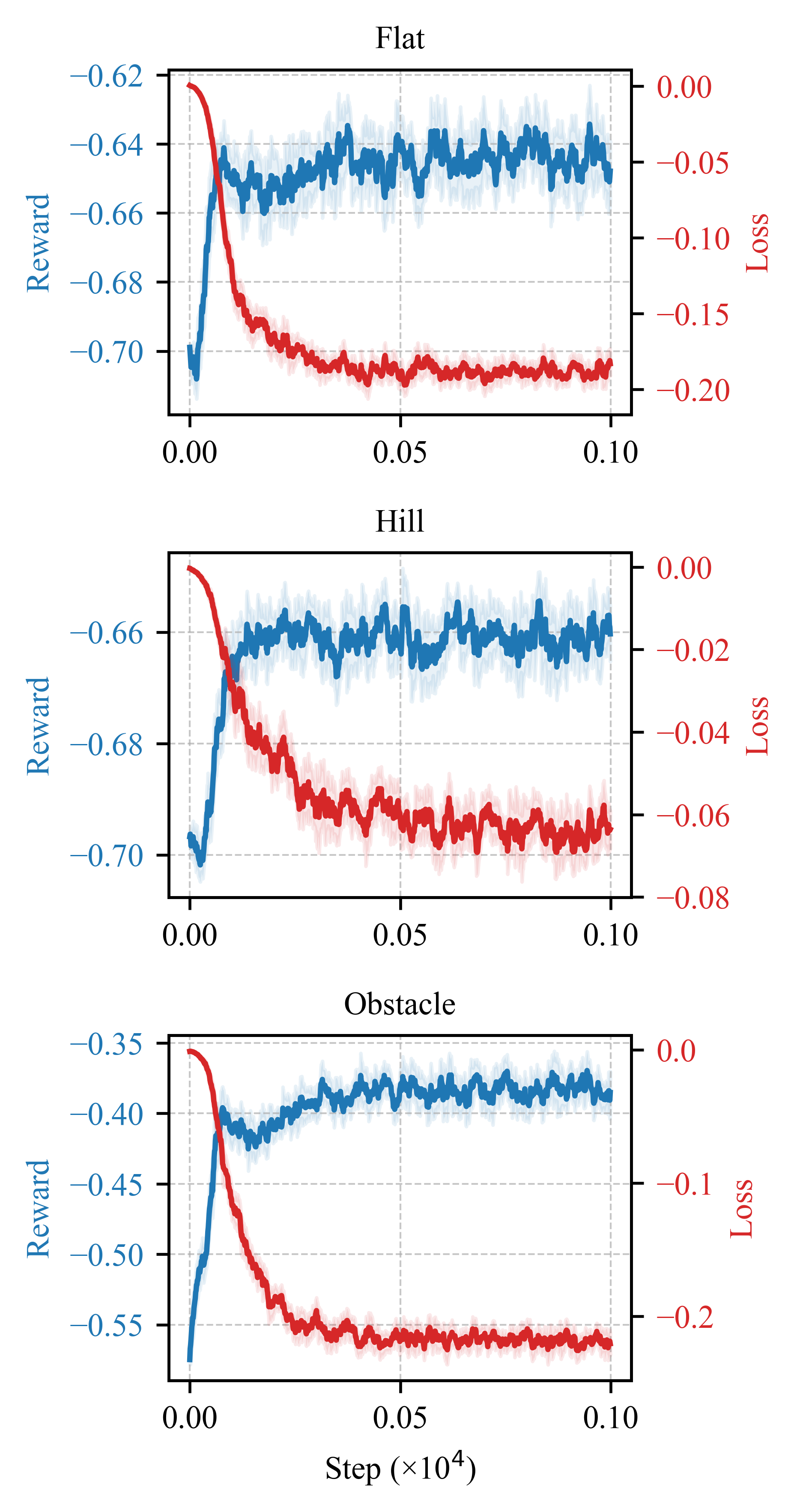}
\caption{}
\label{fig:adv_train}
\end{subfigure}
\caption{Training trajectories of (a) victim agent and (b) AGMR adversarial agent.}
\label{fig:combined_trajectories}
\end{figure}

\subsection{Experimental Setup}
\subsubsection{Environment}
We systematically evaluate quadrupedal locomotion control and adversarial robustness through comprehensive experiments conducted on the RaiSim platform \cite{raisim}, a state-of-the-art physics engine renowned for its high-fidelity robotics simulations. As illustrated in Figure~\ref{fig:raisim}, we employ the ANYmal quadruped robot as our testbed across three meticulously designed tasks:
\begin{itemize}
    \item \textbf{Flat}: The robot moves on a flat and even surface to evaluate its basic locomotion capabilities. 
    
    \item \textbf{Hill}: The robot moves across uneven terrain with varying heights and slopes, designed to test its ability to maintain stability and adapt to unpredictable ground variations.
    
    \item \textbf{Obstacle}: The robot moves in a grid-like obstacle field, where the terrain consists of regularly spaced obstacles that challenge the robot's precision and robustness in foot placement.
\end{itemize}

The state space (34 dimensions) includes body height, body orientation (3), joint angles (12), body linear velocity (3), body angular velocity (3), and joint velocities (12). The action space (12 dimensions) controls the hip, thigh, and calf joints of the front right, front left, rear right, and rear left legs (3 dimensions each). Each episode lasts up to 4 seconds (400 control steps). If the robot falls, defined as any non-foot part touching the ground, the episode ends early. Experiments are conducted on a server with four NVIDIA GeForce RTX 3090 GPUs (24GB each).

\subsubsection{Training}

The victim agent is trained using Proximal Policy Optimization (PPO) \cite{schulman2017proximal}, an on-policy RL algorithm that combines trust region optimization with clipped surrogate objectives. It employs a dual-network architecture consisting of an actor network and a critic network. The network architecture and hyperparameters are summarized in Table~\ref{tab:hyper}. The victim agent's reward function balances energy efficiency and performance through torque penalties and forward velocity incentives:
\begin{equation}
R_{\text{vic}}(\tau,v_x) = \xi \sum_{i=1}^{12} \tau_i^2 + \kappa \min(v_x, 4.0),
\end{equation}
where $\tau_i$ represents the torque of the $i$-th joint, $v_x$ is the forward velocity, $\xi = -4 \times 10^{-5}$ is the energy efficiency coefficient penalizing the sum of squared torques, and $\kappa = 0.3$ is the forward velocity coefficient encouraging locomotion with an upper bound of  4.0 m/s. Figure~\ref{fig:victim_train} shows the training trajectories.

The adversarial agent, trained using the AGMR framework, consists of an adversarial policy and value function. The masking function within the adversarial policy is parameterized by neural networks, as detailed in Table~\ref{tab:hyper}. The adversarial reward is defined as the negative of the victim agent’s reward to minimize artificial effort introduced by reward shaping:
\begin{equation}
R_{\text{adv}} = - R_{\text{vic}}.
\end{equation}
The adversarial agent's training spans $2\times10^3$ steps, with the trajectory shown in Figure~\ref{fig:adv_train}.

\subsubsection{Baselines}
To evaluate the performance of AGMR, we compare it with a wide range of existing adversarial attack methods. For a fair comparison, we set the perturbation budget for all methods to $\epsilon = 0.125$. The methods compared include:

\begin{itemize}
    \item \textbf{Random Attack}: A baseline applying uniform random noise as perturbations, serving as a naive benchmark.
    
    \item \textbf{FGSM} \cite{goodfellow2014explaining}: A single-step gradient-based attack designed to maximize loss with minimal computation.
    
    \item \textbf{DI\textsuperscript{2}-FGSM} \cite{xie2019improving}: An FGSM extension introducing input transformations (e.g., resizing, padding) to enhance transferability.
    
    \item \textbf{MI-FGSM} \cite{dong2018boosting}: An iterative FGSM variant leveraging momentum to stabilize updates and improve transferability.
    
    \item \textbf{NI-FGSM} \cite{lin2019nesterov}: Builds on MI-FGSM by integrating Nesterov accelerated gradients for refined updates.
    
    \item \textbf{R+FGSM} \cite{tramer2017ensemble}: An FGSM extension adding random perturbations before gradient-based updates to mitigate local gradient sensitivity.
    
    \item \textbf{PGD} \cite{madry2017towards}: An iterative FGSM extension with projection onto the allowed perturbation space after each step.
    
    \item \textbf{TPGD} \cite{zhang2019theoretically}: A PGD variant replacing cross-entropy loss with KL divergence to improve attack success against robust models.
    
    \item \textbf{EOT-PGD} \cite{liu2018adv}: A PGD variant applying random transformations or model variations during iterations to ensure robustness across distributions.

\end{itemize}

\begin{table*}[]
\centering
\renewcommand{\arraystretch}{1.1}
\begin{tabular}{|c|ccc|ccc|ccc|}
\hline
\multirow{2}{*}{\textbf{Method}} & \multicolumn{3}{c|}{\textbf{Flat}} & \multicolumn{3}{c|}{\textbf{Hill}} & \multicolumn{3}{c|}{\textbf{Obstacle}} \\ \cline{2-10} 
                                  & \textbf{R ↓}                & \textbf{V ↓}                & \phantom{x}\textbf{F ↑}\phantom{x}               & \textbf{R ↓}                & \textbf{V ↓}                & \phantom{x}\textbf{F ↑}\phantom{x}               & \textbf{R ↓}                & \textbf{V ↓}                & \phantom{x}\textbf{F ↑}\phantom{x}               \\ \hline
\textbf{NoAttack}                & 0.975 ± 0.001             & 3.700 ± 0.004             & 0                        & 0.920 ± 0.003             & 3.669 ± 0.005             & 0                        & 0.881 ± 0.011             & 3.578 ± 0.037             & 0                        \\ \hline
\textbf{Random}                  & 0.967 ± 0.003             & 3.693 ± 0.004             & 0                        & 0.903 ± 0.031             & 3.619 ± 0.120             & 1                        & 0.872 ± 0.011             & 3.569 ± 0.033             & 0                        \\ \hline
\textbf{FGSM}                    & 0.932 ± 0.008             & 3.649 ± 0.020             & 0                        & 0.811 ± 0.184             & 3.108 ± 1.082             & \underline{2}                        & 0.820 ± 0.021             & 3.477 ± 0.052             & 0                        \\ \hline
\textbf{DI\textsuperscript{2}-FGSM}                  & \underline{0.777 ± 0.197}             & \underline{2.852 ± 1.142}             & \textbf{4}                        & \underline{0.769 ± 0.207}             & 2.835 ± 0.844             & \underline{2}                        & 0.605 ± 0.421             & \underline{2.691 ± 1.255}             & \underline{2}                        \\ \hline
\textbf{MI-FGSM}                  & 0.928 ± 0.017             & 3.653 ± 0.034             & 0                        & 0.782 ± 0.117             & 2.990 ± 1.159             & 1                        & 0.772 ± 0.126             & 3.056 ± 0.937             & \underline{2}                        \\ \hline
\textbf{NI-FGSM}                  & 0.838 ± 0.240             & 3.302 ± 0.991             & 1                        & 0.784 ± 0.145             & \underline{2.748 ± 0.997}             & \textbf{3}                        & 0.691 ± 0.424             & 3.153 ± 1.048             & 1                        \\ \hline
\textbf{R+FGSM}                   & 0.833 ± 0.220             & 3.258 ± 0.987             & 2                        & 0.815 ± 0.053             & 2.991 ± 0.531             & \underline{2}                        & 0.811 ± 0.017             & 3.461 ± 0.052             & 0                        \\ \hline
\textbf{PGD}                     & 0.831 ± 0.166             & 3.090 ± 1.004             & \underline{3}                        & 0.827 ± 0.090             & 3.159 ± 0.785             & \underline{2}                        & 0.744 ± 0.177             & 2.802 ± 1.136             & 1                        \\ \hline
\textbf{TPGD}                    & 0.801 ± 0.216             & 2.933 ± 1.026             & 2                        & 0.858 ± 0.045             & 3.417 ± 0.360             & 1                        & 0.590 ± 0.468             & 2.812 ± 1.368             & \underline{2}                        \\ \hline
\textbf{EOTPGD}                  & 0.779 ± 0.274             & 3.007 ± 1.254             & \textbf{4}                        & 0.817 ± 0.120             & 3.149 ± 0.818             & \underline{2}                        & \underline{0.588 ± 0.458}             & 2.780 ± 1.338             & \underline{2}                        \\ \hline
\textbf{AGMR (Ours)}                    & \textbf{0.623 ± 0.328}             & \textbf{2.156 ± 1.482}             & \textbf{4}                        & \textbf{0.698 ± 0.181}             & \textbf{2.401 ± 1.178}             & \textbf{3}                        & \textbf{0.486 ± 0.572}             & \textbf{2.311 ± 1.566}             & \textbf{3}                        \\ \hline
\end{tabular}
\caption{Attack performance comparison across tasks, with metrics Reward (R) and Forward Velocity (V) and Fall Count (F). The ↑ indicates higher value is
better, while the ↓ indicates lower value is better.}
\label{tab:comparison}
\end{table*}

\subsection{Comparative Experiments}
Tables~\ref{tab:comparison} present the comparative experimental results of baseline methods across three tasks. Performance is evaluated using three metrics: Reward (R), Forward Velocity (V), and Fall Count (F). For each configuration, we conduct 10 independent episodes and report the mean ± standard deviation.  The best-performing method is boldfaced, while suboptimal methods are underlined. The results demonstrate that the proposed method AGMR, consistently achieves superior performance. For the Flat task, AGMR achieves the lowest Reward (0.623 ± 0.328) and Velocity (2.156 ± 1.482), while maintaining the highest Fall Count (F = 4), indicating its significant ability to disrupt task performance. Similarly, in the Hill task, AGMR exhibits the lowest Reward (0.698 ± 0.181) and Velocity (2.401 ± 1.178), with a competitive Fall Count (F = 3), outperforming other methods. In the Obstacle task, AGMR further highlights its effectiveness, achieving the lowest Reward (0.486 ± 0.572), the lowest Velocity (2.311 ± 1.566), and the highest Fall Count (F = 3), surpassing other method in destabilizing the locomotion.

\begin{figure}[htbp]
\centering
\begin{subfigure}{\linewidth}
    \centering
    \includegraphics[width=1\linewidth]{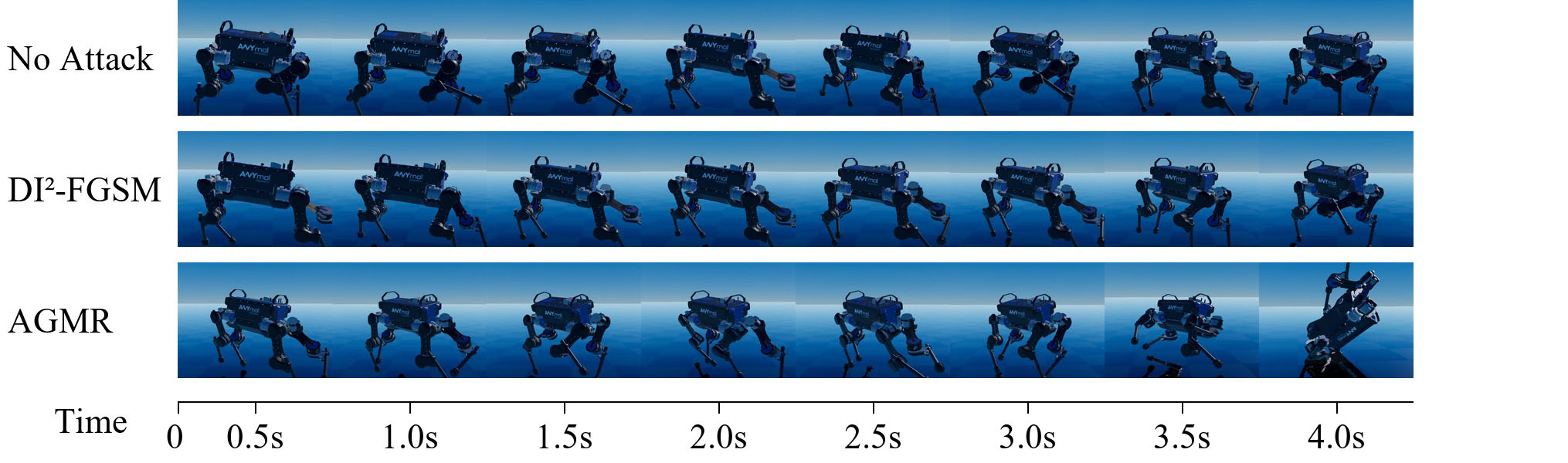}
    \caption{}
    \label{fig:gait1}
\end{subfigure}

\vspace{0.5em} 

\begin{subfigure}{0.95\linewidth}
    \centering
    \includegraphics[width=\linewidth]{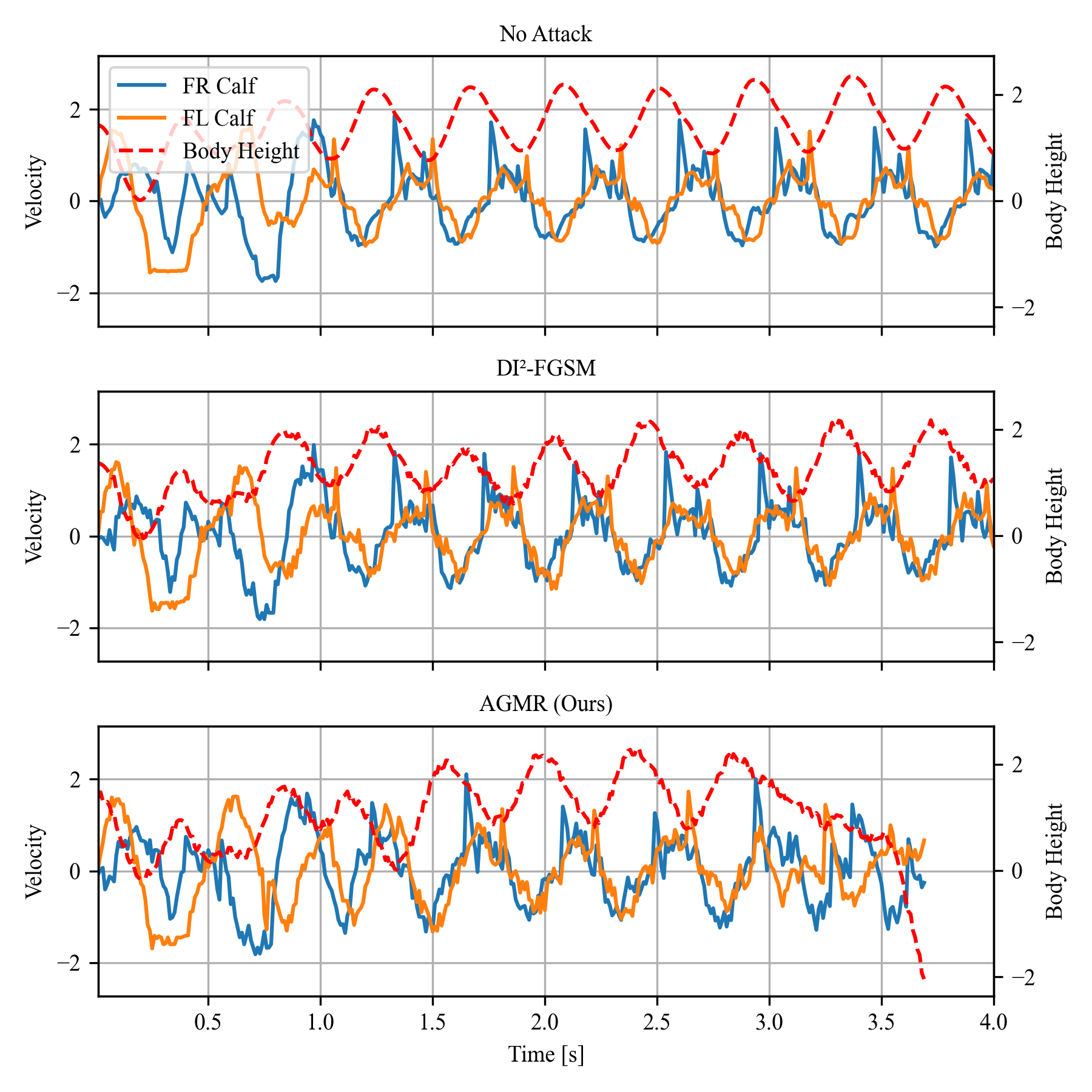}
    \caption{}
    \label{fig:gait2}
\end{subfigure}
    \caption{Locomotion performance under No Attack, DI\textsuperscript{2}-FGSM ($\epsilon = 0.125$), and AGMR ($\epsilon = 0.125$). (a) Visualized motion sequences. (b) Time series of body height and calf trajectories. AGMR induces greater instability under the same perturbation budget.}
\label{fig:gait}
\end{figure}

\subsection{Behavior Analysis}
Figure~\ref{fig:gait1} visualizes motion sequence over time during the flat task under  No Attack, the strongest baseline DI²-FGSM ($\epsilon$ = 0.125), and the proposed AGMR ($\epsilon$ = 0.125). Under the No Attack condition, the robot exhibits smooth and stable movement. DI²-FGSM introduces mild instability, while AGMR causes significant disruptions, with the robot's movements becoming increasingly unbalanced and results in a fall.

Figure~\ref{fig:gait2} illustrates the corresponding gait parameters over time, including the body height and the velocities of the front-left (FL) and front-right (FR) calves. Under the No Attack condition, the trajectories maintain smooth, periodic characteristics, reflecting stable locomotion. When subjected to DI²-FGSM, minor perturbations are observed, yet the gait retains its fundamental periodicity and stability. In contrast, under AGMR with an equivalent perturbation budget, significant disruptions occur: both body height and calf joint velocity trajectories exhibit pronounced oscillations, culminating in the robot's collapse. These results intuitively demonstrate that AGMR is highly effective at disrupting the robot's gait, showing its ability to induce critical locomotion failures.



\begin{figure}
    \centering
    \includegraphics[width=0.85\linewidth]{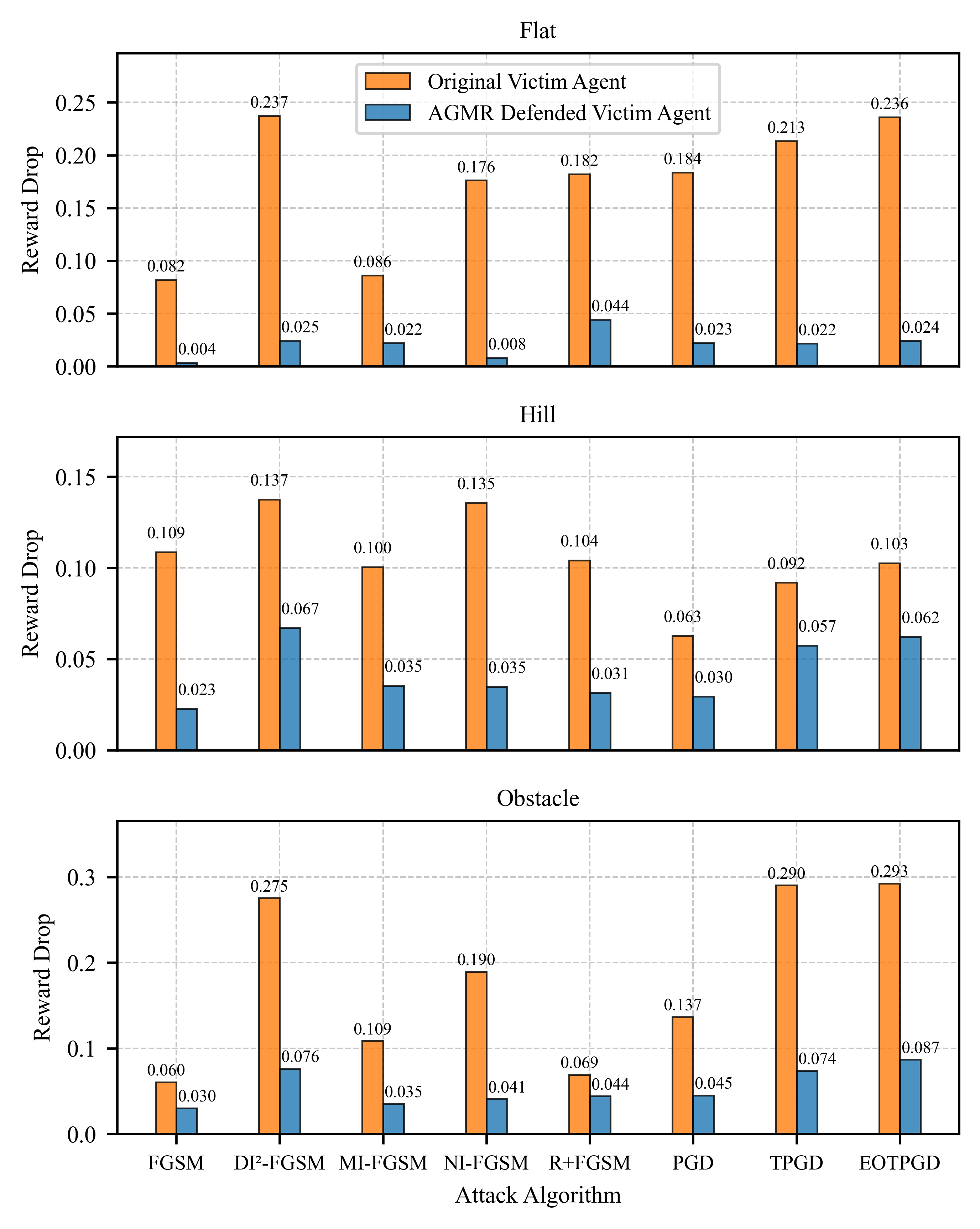}
    \caption{Performance comparison between original and AGMR defended agents under different adversarial attacks.}
    \label{fig:def}
\end{figure}

\subsection{Post-Defense Robustness of AGMR}
Figure~\ref{fig:def} compares the performance of the original victim agent and the AGMR-defended victim agent under various adversarial attacks.
We evaluate the effectiveness of AGMR in enhancing model robustness by exposing the victim agent to AGMR adversarial attacks ($\epsilon$ = 0.125) and training it for $2\times10^2$ steps at a learning rate of $3\times10^{-4}$, while keeping all other hyperparameters constant.
The reward drop is used as the evaluation metric, with lower values indicating better robustness. The results show that the AGMR-defended victim agent demonstrates consistent robustness across diverse attack scenarios and environments, achieving substantial improvements over the original victim agent. In the Flat environment, the AGMR defended victim agent demonstrates strong resilience, with minimal reward drops across all attack methods, consistently below 0.05. In contrast, the original victim agent experiences significant reward degradation, with drops exceeding 0.2 under DI\textsuperscript{2}-FGSM, TPGD, and EOTPGD attack. A similar trend is observed in the Hill environment, where the AGMR-defended victim agent maintains reward drops below 0.05 for most attacks, while the original victim agent shows vulnerability, particularly under DI\textsuperscript{2}-FGSM and NI-FGSM attack, with reward drops significantly surpassing 0.1. In the more challenging Obstacle environment, the original victim agent experiences severe degradation, with reward drops approaching 0.3 under TPGD and EOTPGD attacks. In contrast, the AGMR-defended victim agent achieves significantly lower drops, consistently under 0.1, underscoring its robustness even in complex scenarios.

\begin{figure}
    \centering
    \includegraphics[width=0.85\linewidth]{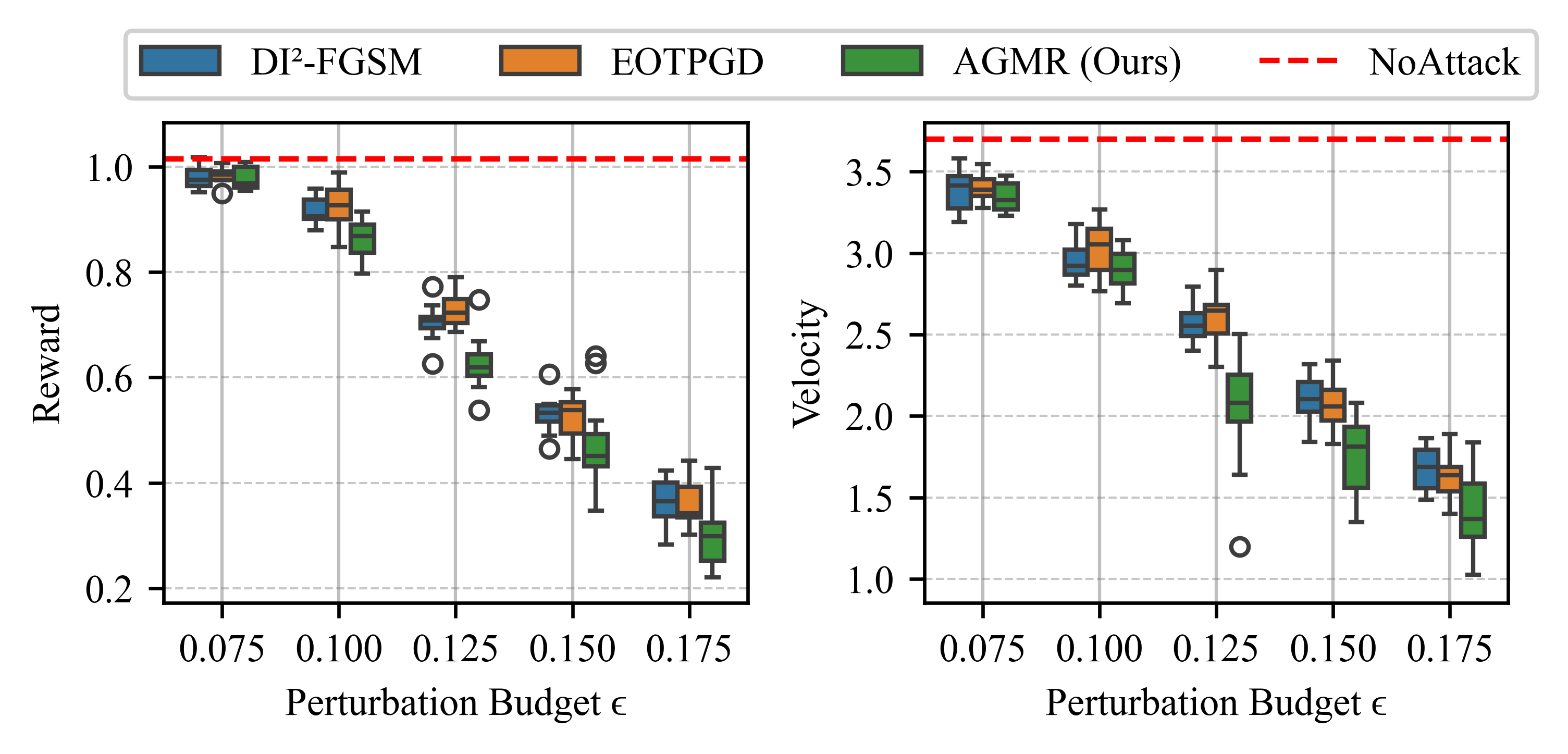}
    \caption{Performance comparison of different attack methods under varying perturbation budgets $\epsilon$.}
    \label{fig:ablation}
\end{figure}

\subsection{Ablation Study}
Figure~\ref{fig:ablation} shows the impact of perturbation budget $\epsilon$ on the attack performance. The perturbation budget $\epsilon$ serves as a crucial hyperparameter that controls the magnitude of adversarial perturbations. A larger $\epsilon$ allows for more substantial modifications to the input observations, potentially leading to more effective attacks, while a smaller $\epsilon$ ensures better imperceptibility. We compare AGMR with two suboptimal methods, DI\textsuperscript{2}-FGSM and EOTPGD, under varying perturbation budgets. The results show AGMR consistently outperforms the baselines across all budgets, with a particularly notable advantage under moderate and high budgets. This improvement is attributed to AGMR’s ability to adaptively allocate perturbation budgets across state dimensions based on their relative importance. By leveraging a gradient-masked reinforcement mechanism, AGMR identifies and exploits critical state dimensions to generate targeted, efficient perturbations that maximize impact within the given budget. In contrast, DI\textsuperscript{2}-FGSM and EOTPGD apply uniform perturbations without considering the varying importance of state dimensions, resulting in less effective attacks.

\section{CONCLUSION}
We propose the Adaptive Gradient-Masked Reinforcement (AGMR) Attack, a white-box adversarial method that integrates DRL with a gradient-based soft masking mechanism to dynamically identify critical state dimensions and optimize adversarial policies. By selectively targeting the most impactful state features, AGMR efficiently disrupts the victim agent's performance. Furthermore, the dynamic adjustment mechanism balances exploration and exploitation, enhancing the adaptability of the attack across diverse tasks and environments. Experimental results demonstrate that AGMR not only outperforms state-of-the-art adversarial attack methods in degrading long-term performance but also improves the robustness of victim agents through adversarial training.   As a potential future direction, we are looking forward to extending our method to improve the performance of various applications such as large language models~\cite{hu2024agentscomerge,lin2024splitlora,hu2021lora,fang2024automated,zhou2025survey,wang2025contemporary}, multi-modal training~\cite{fang2024ic3m,tang2024merit}, and distributed machine learning~\cite{lin2024hierarchical,lyu2023optimal,zhang2024fedac,lin2024efficient,hu2024accelerating,lin2025leo,zhang2024satfed,lin2024fedsn}.

\bibliographystyle{IEEEtran}
\bibliography{reference}

\begin{thebibliography}{10}
\providecommand{\url}[1]{#1}
\csname url@samestyle\endcsname
\providecommand{\newblock}{\relax}
\providecommand{\bibinfo}[2]{#2}
\providecommand{\BIBentrySTDinterwordspacing}{\spaceskip=0pt\relax}
\providecommand{\BIBentryALTinterwordstretchfactor}{4}
\providecommand{\BIBentryALTinterwordspacing}{\spaceskip=\fontdimen2\font plus
\BIBentryALTinterwordstretchfactor\fontdimen3\font minus \fontdimen4\font\relax}
\providecommand{\BIBforeignlanguage}[2]{{%
\expandafter\ifx\csname l@#1\endcsname\relax
\typeout{** WARNING: IEEEtran.bst: No hyphenation pattern has been}%
\typeout{** loaded for the language `#1'. Using the pattern for}%
\typeout{** the default language instead.}%
\else
\language=\csname l@#1\endcsname
\fi
#2}}
\providecommand{\BIBdecl}{\relax}
\BIBdecl

\bibitem{wei2024autonomous}
H.~Wei, B.~Lou, Z.~Zhang, B.~Liang, F.-Y. Wang, and C.~Lv, ``Autonomous navigation for evtol: Review and future perspectives,'' \emph{IEEE Trans. Intell. Veh.}, 2024.

\bibitem{lin2022channel}
Z.~Lin, L.~Wang, J.~Ding, B.~Tan, and S.~Jin, ``Channel power gain estimation for terahertz vehicle-to-infrastructure networks,'' \emph{IEEE Communications Letters}, vol.~27, no.~1, pp. 155--159, 2022.

\bibitem{hu2024toward}
S.~Hu, Z.~Fang, Y.~Deng, X.~Chen, Y.~Fang, and S.~Kwong, ``Toward full-scene domain generalization in multi-agent collaborative bird’s eye view segmentation for connected and autonomous driving,'' \emph{IEEE Transactions on Intelligent Transportation Systems}, 2024.

\bibitem{10135137}
J.~Chen, J.~Cao, Z.~Cheng, and S.~Jiang, ``Towards efficient distributed collision avoidance for heterogeneous mobile robots,'' \emph{IEEE Trans. Mobile Comput.}, vol.~23, no.~5, pp. 3605--3619, 2024.

\bibitem{lin2024adaptsfl}
Z.~Lin, G.~Qu, W.~Wei, X.~Chen, and K.~K. Leung, ``Adaptsfl: Adaptive split federated learning in resource-constrained edge networks,'' \emph{arXiv preprint arXiv:2403.13101}, 2024.

\bibitem{hu2024agentscodriver}
S.~Hu, Z.~Fang, Z.~Fang, Y.~Deng, X.~Chen, and Y.~Fang, ``Agentscodriver: Large language model empowered collaborative driving with lifelong learning,'' \emph{arXiv preprint arXiv:2404.06345}, 2024.

\bibitem{song2023emma}
K.~Song, T.~Ni, L.~Song, and W.~Xu, ``Emma: An accurate, efficient, and multi-modality strategy for autonomous vehicle angle prediction,'' \emph{Intelligent and Converged Networks}, vol.~4, no.~1, pp. 41--49, 2023.

\bibitem{10323097}
S.~T. Atik, A.~S. Chavan, D.~Grosu, and M.~Brocanelli, ``A maintenance-aware approach for sustainable autonomous mobile robot fleet management,'' \emph{IEEE Trans. Mobile Comput.}, vol.~23, no.~6, pp. 7394--7407, 2024.

\bibitem{lin2022tracking}
Z.~Lin, L.~Wang, J.~Ding, Y.~Xu, and B.~Tan, ``Tracking and transmission design in terahertz v2i networks,'' \emph{IEEE Transactions on Wireless Communications}, vol.~22, no.~6, pp. 3586--3598, 2022.

\bibitem{khalil2024advanced}
R.~A. Khalil, Z.~Safelnasr, N.~Yemane, M.~Kedir, A.~Shafiqurrahman, and N.~Saeed, ``Advanced learning technologies for intelligent transportation systems: Prospects and challenges,'' \emph{IEEE Open J. Veh. Technol.}, 2024.

\bibitem{lin2023pushing}
Z.~Lin, G.~Qu, Q.~Chen, X.~Chen, Z.~Chen, and K.~Huang, ``Pushing large language models to the 6g edge: Vision, challenges, and opportunities,'' \emph{arXiv preprint arXiv:2309.16739}, 2023.

\bibitem{fang2024ic3m}
Z.~Fang, Z.~Lin, S.~Hu, H.~Cao, Y.~Deng, X.~Chen, and Y.~Fang, ``Ic3m: In-car multimodal multi-object monitoring for abnormal status of both driver and passengers,'' \emph{arXiv preprint arXiv:2410.02592}, 2024.

\bibitem{lin2022v2i}
Z.~Lin, L.~Wang, J.~Ding, Y.~Xu, and B.~Tan, ``V2i-aided tracking design,'' in \emph{ICC 2022-IEEE International Conference on Communications}, 2022, pp. 291--296.

\bibitem{han2023survey}
D.~Han, B.~Mulyana, V.~Stankovic, and S.~Cheng, ``A survey on deep reinforcement learning algorithms for robotic manipulation,'' \emph{Sensors}, vol.~23, no.~7, p. 3762, 2023.

\bibitem{10100908}
S.~Mohanti, D.~Roy, M.~Eisen, D.~Cavalcanti, and K.~Chowdhury, ``L-norm: Learning and network orchestration at the edge for robot connectivity and mobility in factory floor environments,'' \emph{IEEE Trans. Mobile Comput.}, vol.~23, no.~4, pp. 2898--2914, 2024.

\bibitem{lin2021softgym}
X.~Lin, Y.~Wang, J.~Olkin, and D.~Held, ``Softgym: Benchmarking deep reinforcement learning for deformable object manipulation,'' in \emph{Conf. Robot Learn.}, 2021, pp. 432--448.

\bibitem{duan2025rethinking}
T.~Duan, Z.~Zhang, Z.~Lin, Y.~Gao, L.~Xiong, Y.~Cui, H.~Liang, X.~Chen, H.~Cui, and D.~Huang, ``Rethinking adversarial attacks in reinforcement learning from policy distribution perspective,'' \emph{arXiv preprint arXiv:2501.03562}, 2025.

\bibitem{shi2024distributionally}
L.~Shi and Y.~Chi, ``Distributionally robust model-based offline reinforcement learning with near-optimal sample complexity,'' \emph{J. Mach. Learn. Res.}, vol.~25, no. 200, pp. 1--91, 2024.

\bibitem{yuan2024itpatch}
S.~Yuan, H.~Li, X.~Han, G.~Xu, W.~Jiang, T.~Ni, Q.~Zhao, and Y.~Fang, ``Itpatch: An invisible and triggered physical adversarial patch against traffic sign recognition,'' \emph{arXiv preprint arXiv:2409.12394}, 2024.

\bibitem{schott2024robust}
L.~Schott, J.~Delas, H.~Hajri, E.~Gherbi, R.~Yaich, N.~Boulahia-Cuppens, F.~Cuppens, and S.~Lamprier, ``Robust deep reinforcement learning through adversarial attacks and training: A survey,'' \emph{arXiv preprint arXiv:2403.00420}, 2024.

\bibitem{duan2021advdrop}
R.~Duan, Y.~Chen, D.~Niu, Y.~Yang, A.~K. Qin, and Y.~He, ``Advdrop: Adversarial attack to dnns by dropping information,'' in \emph{Proc. IEEE/CVF Int. Conf. Comput. Vis.}, 2021, pp. 7506--7515.

\bibitem{chen2024diffusion}
J.~Chen, H.~Chen, K.~Chen, Y.~Zhang, Z.~Zou, and Z.~Shi, ``Diffusion models for imperceptible and transferable adversarial attack,'' \emph{IEEE Trans. Pattern Anal. Mach. Intell.}, 2024.

\bibitem{chen2024content}
Z.~Chen, B.~Li, S.~Wu, K.~Jiang, S.~Ding, and W.~Zhang, ``Content-based unrestricted adversarial attack,'' \emph{Adv. Neural Inf. Process. Syst.}, vol.~36, 2024.

\bibitem{oikarinen2021robust}
T.~Oikarinen, W.~Zhang, A.~Megretski, L.~Daniel, and T.-W. Weng, ``{Robust Deep Reinforcement Learning through Adversarial Loss},'' \emph{Adv. Neural Inf. Process. Syst.}, vol.~34, pp. 26\,156--26\,167, 2021.

\bibitem{hickling2023robust}
T.~Hickling, N.~Aouf, and P.~Spencer, ``Robust adversarial attacks detection based on explainable deep reinforcement learning for uav guidance and planning,'' \emph{IEEE Trans. Intell. Veh.}, 2023.

\bibitem{haydari2021adversarial}
A.~Haydari, M.~Zhang, and C.-N. Chuah, ``Adversarial attacks and defense in deep reinforcement learning (drl)-based traffic signal controllers,'' \emph{IEEE Open J. Intell. Transp. Syst.}, vol.~2, pp. 402--416, 2021.

\bibitem{ji2023safety}
J.~Ji, B.~Zhang, J.~Zhou, X.~Pan, W.~Huang, R.~Sun, Y.~Geng, Y.~Zhong, J.~Dai, and Y.~Yang, ``Safety gymnasium: A unified safe reinforcement learning benchmark,'' \emph{Adv. Neural Inf. Process. Syst.}, vol.~36, 2023.

\bibitem{ZW_TCOM_2024}
Q.~Zeng, Z.~Wang, Y.~Zhou, H.~Wu, L.~Yang, and K.~Huang, ``Knowledge-based ultra-low-latency semantic communications for robotic edge intelligence,'' \emph{IEEE Transactions on Communications}, pp. 1--1, 2024.

\bibitem{wang2024ultralola}
\BIBentryALTinterwordspacing
Z.~Wang, A.~E. Kalør, Y.~Zhou, P.~Popovski, and K.~Huang, ``Ultra-low-latency edge inference for distributed sensing,'' 2024. [Online]. Available: \url{https://arxiv.org/abs/2407.13360}
\BIBentrySTDinterwordspacing

\bibitem{akhtar2021advances}
N.~Akhtar, A.~Mian, N.~Kardan, and M.~Shah, ``{Advances in Adversarial Attacks and Defenses in Computer Vision: A Survey},'' \emph{IEEE Access}, vol.~9, pp. 155\,161--155\,196, 2021.

\bibitem{goodfellow2014explaining}
I.~J. Goodfellow, J.~Shlens, and C.~Szegedy, ``{Explaining and Harnessing Adversarial Examples},'' \emph{arXiv preprint arXiv:1412.6572}, 2014.

\bibitem{madry2017towards}
A.~Madry, ``{Towards Deep Learning Models Resistant to Adversarial Attacks},'' \emph{arXiv preprint arXiv:1706.06083}, 2017.

\bibitem{wong2020fast}
E.~Wong, L.~Rice, and J.~Z. Kolter, ``{Fast Is Better Than Free: Revisiting Adversarial Training},'' \emph{arXiv preprint arXiv:2001.03994}, 2020.

\bibitem{schwinn2023exploring}
L.~Schwinn, R.~Raab, A.~Nguyen, D.~Zanca, and B.~Eskofier, ``{Exploring Misclassifications of Robust Neural Networks To Enhance Adversarial Attacks},'' \emph{Appl. Intell.}, vol.~53, no.~17, pp. 19\,843--19\,859, 2023.

\bibitem{wang2021admix}
X.~Wang, X.~He, J.~Wang, and K.~He, ``{Admix: Enhancing the Transferability of Adversarial Attacks},'' in \emph{Proc. IEEE/CVF Int. Conf. Comput. Vis.}, 2021, pp. 16\,158--16\,167.

\bibitem{dong2018boosting}
Y.~Dong, F.~Liao, T.~Pang, H.~Su, J.~Zhu, X.~Hu, and J.~Li, ``{Boosting Adversarial Attacks with Momentum},'' in \emph{Proc. IEEE Conf. Comput. Vis. Pattern Recognit.}, 2018, pp. 9185--9193.

\bibitem{xie2019improving}
C.~Xie, Z.~Zhang, Y.~Zhou, S.~Bai, J.~Wang, Z.~Ren, and A.~L. Yuille, ``{Improving Transferability of Adversarial Examples with Input Diversity},'' in \emph{Proc. IEEE/CVF Conf. Comput. Vis. Pattern Recognit.}, 2019, pp. 2730--2739.

\bibitem{dong2019evading}
Y.~Dong, T.~Pang, H.~Su, and J.~Zhu, ``{Evading Defenses to Transferable Adversarial Examples by Translation-invariant Attacks},'' in \emph{Proc. IEEE/CVF Conf. Comput. Vis. Pattern Recognit.}, 2019, pp. 4312--4321.

\bibitem{lin2019nesterov}
J.~Lin, C.~Song, K.~He, L.~Wang, and J.~E. Hopcroft, ``{Nesterov Accelerated Gradient and Scale Invariance for Adversarial Attacks},'' \emph{arXiv preprint arXiv:1908.06281}, 2019.

\bibitem{ilahi2021challenges}
I.~Ilahi, M.~Usama, J.~Qadir, M.~U. Janjua, A.~Al-Fuqaha, D.~T. Hoang, and D.~Niyato, ``{Challenges and Countermeasures for Adversarial Attacks on Deep Reinforcement Learning},'' \emph{IEEE Trans. Artif. Intell.}, vol.~3, no.~2, pp. 90--109, 2021.

\bibitem{huang2017adversarial}
S.~Huang, N.~Papernot, I.~Goodfellow, Y.~Duan, and P.~Abbeel, ``{Adversarial Attacks on Neural Network Policies},'' \emph{arXiv preprint arXiv:1702.02284}, 2017.

\bibitem{pattanaik2017robust}
A.~Pattanaik, Z.~Tang, S.~Liu, G.~Bommannan, and G.~Chowdhary, ``{Robust Deep Reinforcement Learning with Adversarial Attacks},'' \emph{arXiv preprint arXiv:1712.03632}, 2017.

\bibitem{lin2017tactics}
Y.-C. Lin, Z.-W. Hong, Y.-H. Liao, M.-L. Shih, M.-Y. Liu, and M.~Sun, ``{Tactics of Adversarial Attack on Deep Reinforcement Learning Agents},'' \emph{arXiv preprint arXiv:1703.06748}, 2017.

\bibitem{weng2019toward}
T.-W. Weng, K.~D. Dvijotham, J.~Uesato, K.~Xiao, S.~Gowal, R.~Stanforth, and P.~Kohli, ``{Toward Evaluating Robustness of Deep Reinforcement Learning with Continuous Control},'' in \emph{Int. Conf. Learn. Represent.}, 2019.

\bibitem{zhang2020robust}
H.~Zhang, H.~Chen, C.~Xiao, B.~Li, M.~Liu, D.~Boning, and C.-J. Hsieh, ``{Robust Deep Reinforcement Learning against Adversarial Perturbations on State Observations},'' \emph{Adv. Neural Inf. Process. Syst.}, vol.~33, pp. 21\,024--21\,037, 2020.

\bibitem{sutton2018reinforcement}
R.~S. Sutton, ``Reinforcement learning: An introduction,'' \emph{Bradford Book}, 2018.

\bibitem{mnih2013playing}
V.~Mnih, ``Playing atari with deep reinforcement learning,'' \emph{arXiv preprint arXiv:1312.5602}, 2013.

\bibitem{achiam2017constrained}
J.~Achiam, D.~Held, A.~Tamar, and P.~Abbeel, ``{Constrained Policy Optimization},'' in \emph{Int. Conf. Mach. Learn.}, 2017, pp. 22--31.

\bibitem{chakraborty2018adversarial}
A.~Chakraborty, M.~Alam, V.~Dey, A.~Chattopadhyay, and D.~Mukhopadhyay, ``Adversarial attacks and defences: A survey,'' \emph{arXiv preprint arXiv:1810.00069}, 2018.

\bibitem{bellemare2013arcade}
M.~G. Bellemare, Y.~Naddaf, J.~Veness, and M.~Bowling, ``The arcade learning environment: An evaluation platform for general agents,'' \emph{J. Artif. Intell. Res.}, vol.~47, pp. 253--279, 2013.

\bibitem{schulman2015high}
J.~Schulman, P.~Moritz, S.~Levine, M.~Jordan, and P.~Abbeel, ``{High-dimensional Continuous Control Using Generalized Advantage Estimation},'' \emph{arXiv preprint arXiv:1506.02438}, 2015.

\bibitem{raisim}
\BIBentryALTinterwordspacing
J.~Hwangbo, J.~Lee, and M.~Hutter, ``Per-contact iteration method for solving contact dynamics,'' \emph{IEEE Robot. Autom. Lett.}, vol.~3, no.~2, pp. 895--902, 2018. [Online]. Available: \url{www.raisim.com}
\BIBentrySTDinterwordspacing

\bibitem{schulman2017proximal}
J.~Schulman, F.~Wolski, P.~Dhariwal, A.~Radford, and O.~Klimov, ``Proximal policy optimization algorithms,'' \emph{arXiv preprint arXiv:1707.06347}, 2017.

\bibitem{tramer2017ensemble}
F.~Tram{\`e}r, A.~Kurakin, N.~Papernot, I.~Goodfellow, D.~Boneh, and P.~McDaniel, ``{Ensemble Adversarial Training: Attacks and Defenses},'' \emph{arXiv preprint arXiv:1705.07204}, 2017.

\bibitem{zhang2019theoretically}
H.~Zhang, Y.~Yu, J.~Jiao, E.~Xing, L.~El~Ghaoui, and M.~Jordan, ``{Theoretically Principled Trade-off Between Robustness and Accuracy},'' in \emph{Int. Conf. Mach. Learn.}, 2019, pp. 7472--7482.

\bibitem{liu2018adv}
X.~Liu, Y.~Li, C.~Wu, and C.-J. Hsieh, ``{Adv-bnn: Improved Adversarial Defense through Robust Bayesian Neural Network},'' \emph{arXiv preprint arXiv:1810.01279}, 2018.

\bibitem{hu2024agentscomerge}
S.~Hu, Z.~Fang, Z.~Fang, Y.~Deng, X.~Chen, Y.~Fang, and S.~Kwong, ``Agentscomerge: Large language model empowered collaborative decision making for ramp merging,'' \emph{arXiv preprint arXiv:2408.03624}, 2024.

\bibitem{lin2024splitlora}
Z.~Lin, X.~Hu, Y.~Zhang, Z.~Chen, Z.~Fang, X.~Chen, A.~Li, P.~Vepakomma, and Y.~Gao, ``Splitlora: A split parameter-efficient fine-tuning framework for large language models,'' \emph{arXiv preprint arXiv:2407.00952}, 2024.

\bibitem{hu2021lora}
E.~J. Hu, Y.~Shen, P.~Wallis, Z.~Allen-Zhu, Y.~Li, S.~Wang, L.~Wang, and W.~Chen, ``Lora: Low-rank adaptation of large language models,'' \emph{arXiv preprint arXiv:2106.09685}, 2021.

\bibitem{fang2024automated}
Z.~Fang, Z.~Lin, Z.~Chen, X.~Chen, Y.~Gao, and Y.~Fang, ``Automated federated pipeline for parameter-efficient fine-tuning of large language models,'' \emph{arXiv preprint arXiv:2404.06448}, 2024.

\bibitem{zhou2025survey}
Y.~Zhou, T.~Ni, W.-B. Lee, and Q.~Zhao, ``A survey on backdoor threats in large language models (llms): Attacks, defenses, and evaluations,'' \emph{arXiv preprint arXiv:2502.05224}, 2025.

\bibitem{wang2025contemporary}
J.~Wang, T.~Ni, W.-B. Lee, and Q.~Zhao, ``A contemporary survey of large language model assisted program analysis,'' \emph{arXiv preprint arXiv:2502.18474}, 2025.

\bibitem{tang2024merit}
Y.~Tang, Z.~Chen, A.~Li, T.~Zheng, Z.~Lin, J.~Xu, P.~Lv, Z.~Sun, and Y.~Gao, ``Merit: Multimodal wearable vital sign waveform monitoring,'' \emph{arXiv preprint arXiv:2410.00392}, 2024.

\bibitem{lin2024hierarchical}
Z.~Lin, W.~Wei, Z.~Chen, C.-T. Lam, X.~Chen, Y.~Gao, and J.~Luo, ``Hierarchical split federated learning: Convergence analysis and system optimization,'' \emph{arXiv preprint arXiv:2412.07197}, 2024.

\bibitem{lyu2023optimal}
S.~Lyu, Z.~Lin, G.~Qu, X.~Chen, X.~Huang, and P.~Li, ``Optimal resource allocation for u-shaped parallel split learning,'' in \emph{2023 IEEE Globecom Workshops (GC Wkshps)}, 2023, pp. 197--202.

\bibitem{zhang2024fedac}
Y.~Zhang, H.~Chen, Z.~Lin, Z.~Chen, and J.~Zhao, ``Fedac: A adaptive clustered federated learning framework for heterogeneous data,'' \emph{arXiv preprint arXiv:2403.16460}, 2024.

\bibitem{lin2024efficient}
Z.~Lin, G.~Zhu, Y.~Deng, X.~Chen, Y.~Gao, K.~Huang, and Y.~Fang, ``Efficient parallel split learning over resource-constrained wireless edge networks,'' \emph{IEEE Transactions on Mobile Computing}, 2024.

\bibitem{hu2024accelerating}
M.~Hu, J.~Zhang, X.~Wang, S.~Liu, and Z.~Lin, ``Accelerating federated learning with model segmentation for edge networks,'' \emph{IEEE Transactions on Green Communications and Networking}, 2024.

\bibitem{lin2025leo}
Z.~Lin, Y.~Zhang, Z.~Chen, Z.~Fang, C.~Wu, X.~Chen, Y.~Gao, and J.~Luo, ``Leo-split: A semi-supervised split learning framework over leo satellite networks,'' \emph{arXiv preprint arXiv:2501.01293}, 2025.

\bibitem{zhang2024satfed}
Y.~Zhang, Z.~Lin, Z.~Chen, Z.~Fang, W.~Zhu, X.~Chen, J.~Zhao, and Y.~Gao, ``Satfed: A resource-efficient leo satellite-assisted heterogeneous federated learning framework,'' \emph{arXiv preprint arXiv:2409.13503}, 2024.

\bibitem{lin2024fedsn}
Z.~Lin, Z.~Chen, Z.~Fang, X.~Chen, X.~Wang, and Y.~Gao, ``Fedsn: A federated learning framework over heterogeneous leo satellite networks,'' \emph{IEEE Transactions on Mobile Computing}, 2024.

\end{thebibliography}

\end{document}